\documentclass[a4paper, 10pt, conference]{ieeeconf}

\usepackage{FG2027}

\FGfinalcopy %

\IEEEoverridecommandlockouts  %
\usepackage{graphicx}
\graphicspath{{figs/}}

\usepackage{times}
\usepackage{amsmath}
\usepackage{amssymb}
\usepackage{booktabs}
\usepackage{multirow}
\usepackage[dvipsnames]{xcolor}
\usepackage{comment}
\usepackage{capt-of}  %
\makeatletter
\let\NAT@parse\undefined
\makeatother
\usepackage[numbers,sort&compress]{natbib}
\usepackage[hyphens]{url}

\usepackage{xr-hyper}

\definecolor{fgblue}{rgb}{0.21,0.49,0.74}
\usepackage[pagebackref,breaklinks,colorlinks,allcolors=fgblue]{hyperref}
\usepackage[capitalize]{cleveref}
\crefname{section}{Sec.}{Secs.}
\Crefname{section}{Section}{Sections}
\crefname{subsection}{Sec.}{Secs.}
\Crefname{subsection}{Section}{Sections}
\crefname{table}{Tab.}{Tabs.}
\Crefname{table}{Table}{Tables}
\crefname{figure}{Fig.}{Figs.}
\Crefname{figure}{Figure}{Figures}

\def\FGPaperID{*****}

\title{\LARGE \bf
Parsed Text Understates Vision--Language Models:\\
From Below to Above Expert Agreement on Fine-Grained Facial Emotion
}

\author{\parbox{16cm}{\centering
    {\large Tobias Hallmen$^1$, Fabian Deuser$^2$, Robin-Nico Kampa$^2$, Norbert Oswald$^2$ and Elisabeth Andr\'e$^1$}\\
    {\normalsize
    $^1$ Chair for Human-Centered Artificial Intelligence, University of Augsburg, Germany\\
    $^2$ Institute for Distributed Intelligent Systems, University of the Bundeswehr Munich, Germany\\
    \texttt{\{tobias.hallmen,elisabeth.andre\}@uni-a.de}, \texttt{\{fabian.deuser,robin-nico.kampa,norbert.oswald\}@unibw.de}}}
}

\begin{document}

\ifFGfinal
\thispagestyle{empty}
\pagestyle{empty}
\else
\author{Anonymous FG2027 submission\\ Paper ID \FGPaperID \\}
\pagestyle{plain}
\fi
\maketitle
{\let\thefootnote\relax\footnotetext{Corresponding author: Tobias Hallmen (tobias.hallmen@uni-a.de).}}{\let\thefootnote\relax\footnotetext{This work has been submitted to the IEEE for possible publication. Copyright may be transferred without notice, after which this version may no longer be accessible.}}

\begin{abstract}
EmoNet-Face-HQ is a benchmark of 2,500 generated portraits, each rated by four psychology experts on 40 emotion categories. It concludes that vision--language models (VLMs) on average offer no dependable improvement over random guessing and that a fine-tuned model is required. The benchmark reads each model's answer from generated text, and its zero-shot prompt demands a fixed number of emotions per face. Its own results already show that this protocol can decide the outcome: one VLM moves from chance to close to expert agreement when only the prompt changes. We keep the images, expert ratings and metric and change only how the answer is asked for and read: one yes/no question per category, scored by the probability of ``yes''. Scored from parsed text on the five categories experts agree on most, eleven open-weight VLMs average a quadratic-weighted $\kappa$ of 0.425 against individual experts, below the 0.468 at which two experts agree. Scored from yes-probabilities, the same models average 0.553, and all eleven agree with an individual expert better than a second expert does. The VLMs' mean is level with the benchmark's fine-tuned model (0.555), and three of them significantly exceed it without any training. Only one exceeds a stricter reference, the mean of three experts scored against the fourth (0.563). Verdicts that VLMs fail at graded facial affect ratings may therefore describe the readout more than the model. A fine-tuned model should have to beat a pretrained VLM read this way before it is called necessary, and the benchmark's own does not clearly do so.
\end{abstract}

\section{Introduction}
\label{sec:intro}
Systems that respond to a person's feelings, such as therapy aids or social robots, need to interpret more than the six or seven basic emotions that most benchmarks identify. Vision-language models (VLMs) are an obvious candidate for this task because they are already used not only to describe images, but also to rate them. For example, they can be used to judge image quality~\cite{wu2024qalign} or how well an image matches a caption~\cite{lin2024vqascore}. Whether a VLM can rate a face is decided by benchmarks that compare its ratings with human ones. Such a comparison measures the model through a procedure: the model has to be asked, and its answer has to be read.

Fine-grained labels are hard to obtain, because collecting them on real faces is expensive and raises data-protection questions. EmoNet-Face-HQ~\cite{schuhmann2025emonetface} takes a different route: 2,500 generated portraits, each rated by four psychology experts on 40 emotion categories for intensity, from 0 (absent) to 7 (maximal). Because every image has four raters, the benchmark also shows how far the experts agree with one another, and on most of the 40 categories they agree little. Such disagreement complicates both training a model on the ratings and judging its output against them~\cite{plank2022problem}. We therefore judge a model by whether it agrees with an individual expert as well as a second expert does. We score the five categories on which the experts agree most, fixed before any model was scored (\cref{fig:alpha}), and report all 40 alongside.

The benchmark tests VLMs on these ratings and concludes that, whatever the prompt format, they offer on average no dependable improvement over random guessing~\cite{schuhmann2025emonetface}. They therefore train a task-specific model, Empathic-Insight-Face (EIF), and report it at the level of the experts.  This verdict is difficult to reconcile with the way VLMs perform on other rating tasks. It raises the question of whether the models are failing or the measurements are flawed.

\begin{figure}[t]
\centering
\parbox[c]{0.42\columnwidth}{\centering\includegraphics[height=3.2cm]{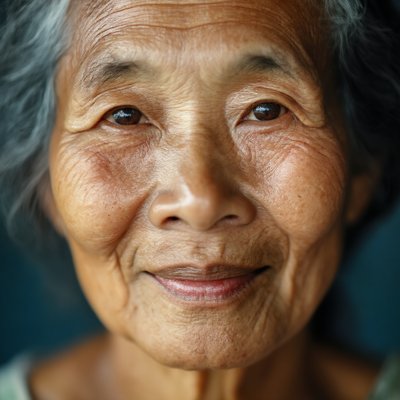}}\hfill
\parbox[c]{0.52\columnwidth}{\footnotesize
\textbf{Reply (Gemma-4-12B-it):}\\
\texttt{Contentment 6, Thankfulness/Gratitude 5, Affection 5, Relief 4, Contemplation 4, Pride 3, Amusement 3, Interest 3, Emotional Numbness 2, Fatigue/Exhaustion 3}\\[2pt]
Ten categories. The other 30 count as 0.}
\\[6pt]
\footnotesize
\begin{tabular}{@{}lcccc@{}}
\toprule
Category & four experts & in reply & $P(\text{yes})$ & pctl. \\
\midrule
Elation                & 0, 1, 4, 4 & no        & 0.011 & 84 \\
Relief                 & 7, 0, 0, 3 & yes (4)   & 0.066 & 89 \\
Thankfulness/Gratitude & 0, 2, 5, 2 & yes (5)   & 0.763 & 92 \\
\bottomrule
\end{tabular}
\caption{One image, read two ways. The written reply names ten of 40 categories and omits Elation, which three of four experts rate as present. The same model's $P(\text{yes})$ for Elation is small but higher than on 84\% of the 2{,}500 images (pctl.), and this rank is what the quantile mapping places on the 0--7 scale. The experts themselves disagree on Relief.}
\label{fig:teaser}
\end{figure}

Figure~\ref{fig:teaser} shows one place where the measurement can go wrong. The benchmark asks each model to write its ratings as text, and its zero-shot prompt demands exactly 25 emotions per face. A model that sees three emotions has to invent 22, and each invented rating is scored against the experts' zeros. A model that ignores the demand runs into a different limit. In Fig.~\ref{fig:teaser} it names ten emotions and omits Elation, which three of the four experts rate as present. A written list cannot say how close the model came to naming Elation; its probability of ``yes'' for Elation is higher than on 84\% of the images. The same figure also shows the experts disagreeing: one rates Relief at 7, two rate it absent.

The benchmark's released results contain a direct test. Gemini 2.5 Flash was run on the same images under two of its prompts. Under the prompt that demands 25 emotions it agrees with the experts no better than the random baseline. Under a second prompt without that demand it reaches a weighted $\kappa$ of 0.15 over all 40 categories, where pairs of experts reach 0.20 on average~\cite{schuhmann2025emonetface}. Model and images are the same; only the prompt changed.

The rating methods cited above do not parse text either: they score a VLM by the probability it assigns to an answer token (\cref{sec:related}). Whether this readout changes the verdict of a published benchmark has not been tested. We test it on EmoNet-Face-HQ with eleven open-weight VLMs. Images, expert ratings and metric stay as the benchmark defines them. Instead of one written list of ratings, each model answers one yes/no question per category, and we read the probability it assigns to ``yes''.

Read from written text, the eleven models agree with an individual expert less well on average than two experts agree with each other. Read from probabilities, every model agrees with an individual expert better than a second expert does, and several exceed EIF without any training. About half of this gain comes from asking about one category at a time, the other half from reading a probability instead of a printed answer. Whether the question asks for yes/no or for a digit from 0 to 7 makes no difference. Fine-tuning the models on the same yes/no questions leaves the score unchanged for six of eight models. On real photographs of posed expressions (FACES~\cite{ebner2010faces}) the gain is small and not consistent across models.

We make three contributions. (i) A re-measurement of EmoNet-Face-HQ that changes only the readout, together with a within-model control that the benchmark's own predictions contain. (ii) A decomposition of the gain into question format and readout, with controls for every choice our protocol adds. (iii) A scoring protocol for graded labels, with checks that flag when its numbers are not measurements, and three recommendations for benchmarks that score VLMs on graded affect labels.

Code and all per-image predictions are released at \url{https://github.com/saveli/emonet-readout}.

\section{Related Work}
\label{sec:related}

\textbf{Fine-grained affect benchmarks.}
In-the-wild facial-affect sets label web images with the basic categories~\cite{mollahosseini2019affectnet,li2017rafdb,ekman1971constants}; posed corpora (FACES~\cite{ebner2010faces}, RaFD~\cite{langner2010rafd}) trade scale for identity and validation ratings. Finer taxonomies are motivated by evidence that experience is not well described by six categories~\cite{cowen2017selfreport,cowen2020face} and contested by evidence that facial configurations do not map onto categories reliably enough to support inference~\cite{barrett2019reconsidered}. That is precisely the regime a $40$-way taxonomy must be evaluated in. EmoNet-Face~\cite{schuhmann2025emonetface} sits at the extreme, with $40$ categories, synthetic balanced imagery and multi-expert ratings. Like comparable evaluations~\cite{lian2024gpt4v}, it scores VLMs by parsing generated text.

\textbf{Prompt and scoring sensitivity in LLM and VLM evaluation.}
Benchmark scores depend on how a model is queried, down to prompt format and option order~\cite{sclar2024quantifying,mizrahi2024state,zheng2024notrobust,pezeshkpour2024order}, enough to move leaderboards by eight positions~\cite{alzahrani2024benchmarks} and to motivate CircularEval~\cite{liu2024mmbench}. Closer to our concern is \emph{how the answer is scored}. Likelihood scoring, symbol scoring, prompting and first-token probabilities are each a different measurement~\cite{holtzman2021surface,robinson2023leveraging,hu2023prompting,wang2024firsttoken}. One model's MMLU~\cite{hendrycks2021mmlu} score differs under cloze scoring, a harness~\cite{gao2023lmeh} and generation~\cite{liang2023helm,fourrier2023leaderboard}. Probabilities carry biases that must be corrected before use~\cite{zhao2021calibrate}. These studies measure how far scores move with the protocol on multiple-choice text benchmarks. On graded face ratings, we show that it moves VLMs from below to above expert--expert agreement.

\textbf{Scoring from token probabilities.}
Reading a score from answer-token probabilities, and not from generated text, is established practice in several fields. VQAScore~\cite{lin2024vqascore} scores image--text alignment as a VLM's probability of ``Yes'' to whether the image shows the text, the closest precedent for probability-based visual scoring. Closest to our query, Atabuzzaman et al.~\cite{atabuzzaman2025zeroshot} classify fine-grained categories by the probability of \texttt{Yes} to a per-class binary question. The same scoring rule rates low-level vision quality~\cite{wu2024qbench,wu2024qalign} and serves LLM judges~\cite{liu2023geval}. In affect, a trained probe on hidden states halves the error of generated answers~\cite{wen2026beyond}, and concurrent speech work separates emitted answers from option logits~\cite{peng2026separating}. The move is disputed: probabilities can misalign with what a model would say~\cite{lyu2024beyond}. Recent decision models make this readout their interface: Jev and its visual variants answer structured choice questions with a probability per candidate, read from the language-model head of a backbone adapted for that use~\cite{typesafe2026jev,yu2026visual,zhou2026text}. Each of these works builds its own evaluation around the probability score. We train nothing, use off-the-shelf models (training on the same questions leaves the score unchanged for six of eight models, \cref{sec:app-training}), and apply the same scoring rule to an evaluation already published, on a $0$--$7$ rating over $40$ co-occurring categories. The difference between the two scoring rules then measures the protocol, not the models, and a control asking for one $0$--$7$ digit per category separates two explanations the literature leaves entangled: asking a per-category question and reading a probability are different interventions (\cref{sec:format}).

\textbf{Reliability, agreement and label noise.}
Agreement statistics, such as Krippendorff's $\alpha$~\cite{krippendorff2018content,hayes2007answering} and weighted $\kappa$~\cite{cohen1968weighted} read against conventional thresholds~\cite{landis1977measurement}, normally certify a corpus as fit to publish. Following Schuhmann et al.~\cite{schuhmann2025emonetface}, we use them as the scale for performance. Agreement between humans is a reference, not a ceiling~\cite{reidsma2008reliability,richie2022ceiling}. Disagreement is often signal, not defect, and metrics against one aggregated label overstate what any individual annotator would endorse~\cite{pavlick2019inherent,nie2020collective,aroyo2015truth,plank2022problem,gordon2021disagreement}. In vision, label errors and re-annotation reorder rankings~\cite{northcutt2021pervasive,beyer2020imagenet,recht2019imagenet}. $\kappa$ also depends on prevalence: when positive labels are rare, high raw agreement can coexist with low $\kappa$~\cite{feinstein1990kappa}. On imbalanced facial-action data, the choice of metric changes how systems compare~\cite{jeni2013imbalanced}. We therefore score models exactly as experts score each other, per category against individual raters (\cref{sec:data}), and separate detection from intensity (\cref{sec:readout}).

\section{The Benchmark's Protocol and Why It Can Fail}
\label{sec:benchmark}

\textbf{Data and reference.}
\label{sec:data}
EmoNet-Face-HQ~\cite{schuhmann2025emonetface} contains $2{,}500$ synthetic portraits generated with MidJourney~v6~\cite{midjourney2024v6} and Flux-Dev and -Pro~\cite{blackforest2024flux}. Four psychology experts rate each image on $40$ emotion categories, on a $0$--$7$ scale where $0$ marks absence and $1$--$7$ presence with intensity. The benchmark's metric is Cohen's $\kappa$ with quadratic weights ($\kappa_w$)~\cite{cohen1968weighted}, which penalizes a disagreement by the square of its distance on that scale. We call the $\kappa_w$ between two individual experts, averaged over the six expert pairs and over categories, \emph{expert--expert agreement}. A model is scored the same way, against each individual expert and not against their mean, so the comparison is like for like. A model at this level agrees with an individual expert as well as a second expert does. It does not match the experts' consensus.

Expert--expert agreement is $0.202$ over all $40$ categories, matching the benchmark's published pairwise $\kappa_w$ of $0.20$. It is $0.468$ on the five categories whose reliability (Krippendorff's $\alpha$~\cite{krippendorff2018content}) is at least $0.30$: \emph{Elation} ($\alpha = 0.584$), \emph{Amusement} ($0.556$), \emph{Anger} ($0.462$), \emph{Astonishment/Surprise} ($0.436$) and \emph{Thankfulness/Gratitude} ($0.300$). On the other $35$ it is $0.163$ (\cref{fig:alpha,tab:percat}). Agreement is low because the taxonomy is fine: many of the $40$ categories are close in meaning or hard to read from a still face, and at this granularity a static face may not determine the category at all~\cite{barrett2019reconsidered}. \emph{Interest}, for example, has $\alpha = -0.08$. We use the five main categories because only there is the reference reliable enough to tell a model's agreement apart from the experts'. The threshold is close, as the sixth category, \emph{Fear}, has $\alpha = 0.298$, so every result is also reported on the other $35$ and on all $40$ (\cref{sec:robustness}).

\begin{table}[t]
\centering\scriptsize\setlength{\tabcolsep}{2pt}
\begin{tabular}{@{}lcc@{\hspace{8pt}}lcc@{}}
\toprule
System & n/img & $\kappa_w$ & System & n/img & $\kappa_w$ \\
\midrule
EIF Small$^{\dagger}$ & $17.2$ & $0.555$ & GPT-4o ZS & $25.0$ & $0.008$ \\
EIF Large$^{\dagger}$ & $18.7$ & $0.539$ & \emph{Random Baseline} & $40.0$ & $0.008$ \\
Gemini 2.0 Flash ZS & $19.5$ & $0.527$ & Nova Pro ZS & $25.0$ & $0.005$ \\
\emph{experts (reference)} & $8.1$ & $0.468$ & Qwen Plus ZS & $25.0$ & $0.004$ \\
Hume Face* & $31.0$ & $0.463$ & Pixtral Large ZS & $25.0$ & $0.004$ \\
Gemini 2.5 Pro ZS & $4.8$ & $0.444$ & Claude 3.7 MS & $4.7$ & $0.000$ \\
\textbf{Gemini 2.5 Flash MS} & $4.9$ & $\mathbf{0.432}$ & \textbf{Gemini 2.5 Flash ZS} & $25.0$ & $\mathbf{-0.001}$ \\
Gemini 2.0 Flash MS & $4.2$ & $0.385$ & & & \\
\bottomrule
\end{tabular}

\caption{The benchmark's fourteen published systems through our scoring path, five main categories, sorted by $\kappa_w$. n/img: mean nonzero categories per image. ZS/MS: zero-/multi-shot prompt. Bold: the within-model control. $^{\dagger}$Trained on EmoNet-Face. $^{*}$Name as listed by the benchmark. Full table: \cref{tab:published-full}.}
\label{tab:published}
\end{table}

\textbf{What the ratings and the metric measure.}
The ratings record how observers perceive a displayed expression, not what anyone felt. $\kappa_w$ rewards both detecting a category and rating its intensity. Presence is rare: a main category is marked present in only $13\%$ (\emph{Astonishment/Surprise}) to $28\%$ (\emph{Thankfulness/Gratitude}) of expert ratings. At such prevalence $\kappa$ can be low despite high raw agreement (\cref{sec:related}), and $\kappa_w$ becomes mostly a detection score. Two experts agree at $\kappa = 0.388$ on presence alone (\cref{sec:readout}), against $\kappa_w = 0.468$ on the full scale.

The benchmark's remedy is EIF, in a Small and a Large size: $40$ MLP heads over frozen SigLIP2 embeddings~\cite{tschannen2025siglip2}. EIF is pre-trained on the $203{,}201$ images of EmoNet-Face-Big and fine-tuned on EmoNet-Face-Binary, neither of which the HQ experts rated, so EIF never saw the labels it is scored against.

\textbf{Text parsing.}
\label{sec:elicitations}
The benchmark scores a model by having it write its answer. The VLM returns one JSON object for all $40$ categories that maps category names to integer ratings, for example \texttt{\{"Anger":6,"Disgust":2\}}, and a parser extracts them. The zero-shot prompt requires ``exactly $25$ key-value pairs'' and tells the model to omit any category it would rate $0$~\cite{laion2025emonetcode}. A reply with a different count, or with any zero, is discarded, and the image is queried again, up to $20$ times (the \emph{retry loop}). Scoring reinstates the omitted categories at $0$ and compares all $40$ ratings with the expert ratings. The benchmark also publishes a \emph{multi-shot} prompt. It supplies worked examples and asks for $44$ dimensions ($40$ emotions plus Arousal, Valence, Dominance and Emotional Vulnerability), each rated and justified, with no fixed count and no rule to omit zeros.

\textbf{Why it can fail.} Under the zero-shot prompt, the number of emotions a model names is set by the prompt, not by the face (\Cref{sec:intro}). Parsed ratings also take at most eight values and give most images 0, so the rank-based mapping we turn to next meets large ties.

\section{Scoring}

\textbf{Quantile mapping.}
\label{sec:scoring}
$\kappa_w$ needs ratings on the levels $0$--$7$, while $P(\mathrm{yes})$ lies in $[0,1]$, and parsed ratings use the scale differently from model to model. The benchmark reaches the scale by truncating predictions to integers, which would map every $P(\mathrm{yes})$ below $1.0$ to $0$. We instead map a prediction $p$ for category $c$ to $\hat y = G_c^{-1}(\hat F_c(p))$, where $\hat F_c$ is the empirical distribution function of the predictions and $G_c^{-1}$ the quantile function of the pooled individual expert ratings: a prediction at a given rank receives the level the expert ratings hold at that rank. This places both conditions on the $0$--$7$ scale without rewarding or penalizing a predictor for the range it happens to emit. Predictions that tie form one block of ranks, and the block is spread over the levels its ranks cover, in proportion. The score is then the $\kappa_w$ of the expected confusion matrix, which is what a random order within each tie gives on average. Parsed ratings pass through the same mapping, so neither condition gains from it alone. A mapping fitted on all images would see the ratings of the images it scores, and a high score could then come from the mapping and not from the model. We therefore split the images in half, fit the mapping on one half and apply it to the other, and vice versa. Applied to the other half, a prediction takes the levels the fitted mapping gives its value, or the nearest value seen in the fit. The mapped predictions of both halves are then pooled and scored together. Every mapped $\kappa_w$ we report is the mean over $20$ random splits. The in-sample mapping, fitted on all images, is the upper-bound row of \cref{tab:controls}. Each $\kappa_w$ is computed between one condition's mapped predictions and one expert's ratings, then averaged over the four experts and over categories, as for expert--expert agreement.

\textbf{Three-expert reference.}
\label{sec:anchor}
Expert--expert agreement compares a model with one second expert. The mean of several experts is a better annotator than any one of them, so it sets a stricter reference. We average three experts, map the mean onto the pooled ratings of those three, and score it against the fourth. Each expert is held out once, and the four results are averaged. This \emph{three-expert reference} is $\kappa_w = 0.563$ $[0.544, 0.578]$. For this comparison each model is mapped onto the same three experts' ratings and scored against the same held-out expert, so model and panel are scored on identical terms. This lowers the models' mean from $0.553$ to $0.543$.

\textbf{Intervals.}
\label{sec:intervals}
Confidence intervals (CIs) resample images, the independent unit, with $B=500$ replicates. The quantile mapping, with its $20$ splits, is refit inside every replicate, since mapping once and then resampling understates the width. A comparison against a reference recomputes both sides on each resample, so the difference is paired; we call it significant when its $95\%$ percentile interval excludes zero. Beside $\kappa_w$ we report mean average precision (mAP) over categories, counting a category as present when any expert rates it above $0$. It depends only on how a model orders the images, not on the mapping onto eight levels, so it checks that a gain does not come from the mapping.

\section{Evidence in the Benchmark's Own Predictions}
\label{sec:evidence}
\label{sec:published}

The benchmark publishes per-image predictions for its fourteen systems, including EIF and a Random Baseline. Our eleven VLMs were never run by the benchmark, so on their own they cannot show that its systems were mismeasured. We therefore score its fourteen exactly as our own models: through the quantile mapping of \cref{sec:scoring} and with $\kappa_w$ against individual experts (\cref{tab:published}). The mapping is needed here as well, because the fourteen do not share one output scale.

\textbf{A within-model control.}
The contrast of \cref{sec:intro} holds through our scoring path on the five main categories: under the zero-shot prompt Gemini 2.5 Flash names $25$ categories on every image and scores $\kappa_w = -0.001$; under the multi-shot prompt it names $4.93$ (SD $0.34$) and scores $0.432$. For this model, the prompt bundle, with its count demand, examples and output format, accounts for the difference.

\textbf{Constant counts.}
Five API-served VLMs under the zero-shot prompt name $25$ emotions on essentially every image. Hume Face, a commercial facial-expression model that returns a continuous score per category, names $31$ on every image.

\textbf{The split follows the prompt.}
The fourteen models fall into two groups. Seven score between $0.385$ and $0.555$. The other seven, the benchmark's own Random Baseline among them, score $0.008$ or less, and five of the remaining six fall below that baseline. Four of those five ran under the zero-shot prompt. Two zero-shot entries escape the count demand. Gemini 2.0 Flash varies its count with the image (SD $8.51$) and scores $+0.058$ $[+0.044,+0.071]$ above expert--expert agreement (paired image bootstrap, \cref{sec:intervals}). Gemini 2.5 Pro names $4.8$ per image and scores $0.444$.

\textbf{Counter-cases.}
Across models, the prompt is not the only factor. Hume Face also names a constant count, yet it is indistinguishable from expert--expert agreement ($-0.005$ $[-0.023,+0.010]$). Its scores are continuous, and continuous scores rank informatively however many are nonzero. Claude 3.7 scores at chance ($0.000$) under the same multi-shot prompt that lifts Gemini 2.5 Flash. Across models, the model and the kind of output it returns also matter.

\textbf{The mapping does not produce the split.}
Across the fourteen, the quantile mapping of \cref{sec:scoring} moves $\kappa_w$ by $+0.044$ on average, and by $+0.010$ over all but the two largest gains. Those two, Hume Face ($+0.361$) and EIF Large ($+0.138$), belong to systems that never output the $0$--$7$ scale: Hume's predictions lie in $[0, 1.03]$, and EIF's in $[-9.21, 9.83]$. The mapping corrects their scale without adding skill, so EIF is read here at its best. The benchmark's own published scores, computed without any mapping on all 40 categories, split the same seven from the same seven ($0.109$--$0.180$ against at most $0.002$; \cref{tab:published-full}).

The prompt explains the published failures. The next sections show that even a model that escapes the count demand is understated by parsed text, and that P(yes) scoring recovers the difference.

\section{P(yes) Scoring}
\label{sec:method}
\label{sec:protocol}

\textbf{Models.}
\label{sec:models}
We evaluate the eleven open-weight VLMs in \cref{tab:models}, from seven model families and between $1.3$B and $14$B parameters, run locally in bfloat16. Open weights make every run reproducible and give access to the first-token probability the protocol reads, so we did not run the API-served systems of \cref{sec:evidence} (\cref{sec:closed}). All eleven models are scored both ways. Decoding and input resolution follow each model's card; holding either uniform moves the gain by at most $0.005$ (\cref{sec:settings}). \Cref{tab:cohorts} lists which models enter each experiment.

\begin{table}[t]
\centering\small
\begin{tabular}{@{}lrll@{}}
\toprule
Model & Params & Prompt & Decoding \\
\midrule
Gemma-3-4b-it~\cite{gemma2025gemma3}                       & $4.30$B & ours & greedy \\
Gemma-3-12b-it~\cite{gemma2025gemma3}                      & $12.19$B & theirs & greedy \\
Gemma-4-12B-it~\cite{gemma2026gemma4}                      & $11.96$B & theirs & greedy \\
GLM-4.6V-Flash~\cite{zai2025glm46vflash,glmv2025glm45v}                       & $10.29$B & theirs & $T=0.8$ \\
InternVL3\_5-8B~\cite{wang2025internvl35,chen2024internvl}  & $8.53$B & theirs & greedy \\
MiMo-VL-7B-RL~\cite{xiaomi2025mimovl}                      & $8.31$B & ours & $T=0.3$ \\
MiniCPM-V-4.6~\cite{openbmb2026minicpmv46,yu2025minicpmv45}       & $1.30$B & ours & $T=1.0$ \\
Ministral-3-14B~\cite{liu2026ministral3}                   & $13.95$B & theirs & greedy \\
Qwen2.5-VL-3B~\cite{bai2025qwen25vl}                       & $3.75$B & theirs & greedy \\
Qwen3-VL-8B~\cite{bai2025qwen3vl}                          & $8.77$B & theirs & greedy \\
Qwen3.5-9B~\cite{qwen2026qwen35}                           & $9.65$B & theirs & $T=0.7$ \\
\bottomrule
\end{tabular}
\caption{The eleven evaluated VLMs. Parameter counts are the released \texttt{safetensors} totals. Prompt and decoding identify the run behind the reported text-parsing score: \emph{theirs} is the benchmark's zero-shot prompt~\cite{laion2025emonetcode}, \emph{ours} a second prompt without the fixed-count demand (\cref{sec:app-prompts}). P(yes) scoring samples nothing.}
\label{tab:models}
\end{table}

\textbf{The query.}
\label{sec:verification}
For each of the $40$ categories we ask one question, ``Does this face express $X$?''. We take $P(\mathrm{yes})$ at the first answer token as the summed probability of the \texttt{yes} tokens over that of the \texttt{yes} and \texttt{no} tokens, each counted with its case and leading-space variants. Each query is one forward pass, and nothing is sampled. Every category receives a probability whether or not the model would have mentioned it. Every image is therefore covered, and the number of categories a model affirms depends on the image, not on the prompt. Several reasoning models emit a chain-of-thought token first, and disabling thinking does not reliably prevent it. We therefore begin every answer with the prefix \texttt{Answer:}. Text parsing has no such token to fix and gets no prefix. \Cref{sec:prefill} measures this asymmetry.

\textbf{Text parsing as we run it.}
We score two text-parsing prompts, the benchmark's zero-shot prompt and a second prompt that removes the fixed-count demand while keeping sparse JSON ratings (\cref{sec:app-prompts}), because a single phrasing cannot distinguish a protocol effect from a wording effect. We report the better of the two per model, which favors text parsing. Our main results do not use the benchmark's retry loop, because it measures compliance with the $25$-key demand, not what the model sees (\cref{sec:retry}); \cref{tab:controls} adds it as a control. A reply the parser cannot read is excluded from that model's sample, not counted as wrong, which also favors text parsing. The exclusion is per image, not per category: a reply that parses but names three categories is kept and scored with $37$ zeros. The share of images a model's replies cover, its \emph{coverage}, is $2{,}500$ of $2{,}500$ for five models and between $1{,}931$ ($77\%$) and $2{,}498$ for the other six, and \cref{sec:depends} holds it fixed. The count demand explains the published systems of \cref{sec:evidence}. Without the retry loop our eleven models often ignore it (\cref{fig:teaser} shows a ten-category reply), so for them the gain measures the readout itself: what parsed ratings lose by leaving most categories at $0$ and using at most eight levels.

\textbf{Validity checks.}
\label{sec:guards}
\label{sec:gate}
Either scoring rule can return a number that is not a measurement. Neither failure announces itself, so each scoring rule carries a check that we report rather than assume. For P(yes) scoring, the \emph{answer-mass check} reports the \emph{answer mass}, the share of first-token probability held by the \texttt{yes} and \texttt{no} variants. If it is near zero, $P(\mathrm{yes})$ is a ratio between two tokens the model was never going to emit. Without the answer prefix, three reasoning-tuned models do this, at a mean answer mass of at most $2.6\times10^{-7}$. The mean answer mass of every run we report lies in $[0.798, 1.000]$, so no threshold has to be chosen (\cref{sec:app-prefill}). For text parsing on the single-label FACES task (\cref{sec:replication}), a reasoning model truncated while reciting the prompt's option list still leaves a category word in its reply. The parser then reports a high parse rate although the model never reached an answer. Our \emph{truncation check} measures the \emph{single-answer gap}: accuracy over replies naming exactly one category, minus accuracy over all replies. A large gap means the parser is producing the score instead of the model.

\section{Results on Eleven VLMs}
\label{sec:results}

\begin{figure*}[t]
\centering
\includegraphics[width=\linewidth]{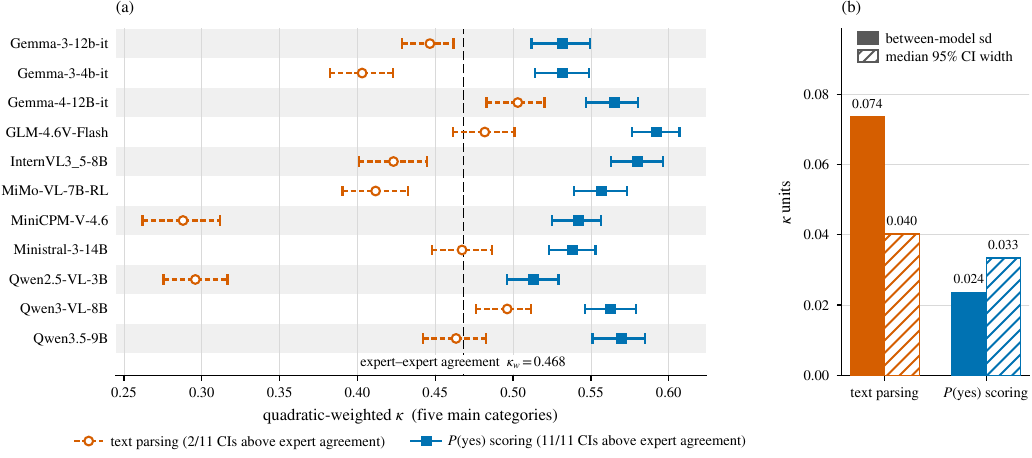}
\caption{\textbf{Under P(yes) scoring every model's interval lies above expert--expert agreement.} Five main categories. (a) $\kappa_w$ with $95\%$ image-bootstrap intervals under text parsing (open circles, dashed) and P(yes) scoring (filled squares), against expert--expert agreement (vertical line). (b) The between-model SD falls while the median CI width also narrows, so P(yes) scoring does not bring the models together by being noisier.}
\label{fig:forest}
\end{figure*}

We compare each model with the three references a reader of the benchmark would use: expert--expert agreement, the benchmark's fine-tuned EIF and the stricter three-expert reference. \Cref{fig:forest} and \cref{tab:eif} give each model under both conditions on the five main categories. With the answer prefix and the quantile mapping, P(yes) scoring raises mean $\kappa_w$ over the eleven models from $0.425$ to $0.553$, by $+0.128$ $[+0.121, +0.133]$.

\begin{table}[t]
\centering\scriptsize\setlength{\tabcolsep}{2.5pt}
\begin{tabular}{lcccccc}
\toprule
 & \multicolumn{3}{c}{$\kappa_w$} & \multicolumn{3}{c}{$P(\mathrm{yes})$ $-$ reference} \\
\cmidrule(lr){2-4}\cmidrule(lr){5-7}
Model & text & matched & P(yes) & 3 experts & EIF S & EIF L \\
\midrule
Gemma-3-12b-it & $0.447$ & $0.436$ & $0.532$ & $-0.040^{*}$ & $-0.023^{*}$ & $-0.007$ \\
Gemma-3-4b-it & $0.403$ & $0.405$ & $0.532$ & $-0.039^{*}$ & $-0.024^{*}$ & $-0.007$ \\
Gemma-4-12B-it & $0.503$ & $0.513$ & $0.565$ & $-0.008$ & $+0.010$ & $+0.026^{*}$ \\
GLM-4.6V-Flash & $0.482$ & $0.550$ & $0.592$ & $+0.018^{*}$ & $+0.037^{*}$ & $+0.053^{*}$ \\
InternVL3\_5-8B & $0.423$ & $0.455$ & $0.580$ & $+0.008$ & $+0.025^{*}$ & $+0.041^{*}$ \\
MiMo-VL-7B-RL & $0.412$ & $0.418$ & $0.557$ & $-0.015^{*}$ & $+0.001$ & $+0.018^{*}$ \\
MiniCPM-V-4.6 & $0.288$ & $0.496$ & $0.542$ & $-0.031^{*}$ & $-0.014^{*}$ & $+0.003$ \\
Ministral-3-14B & $0.467$ & $0.489$ & $0.538$ & $-0.033^{*}$ & $-0.017^{*}$ & $-0.001$ \\
Qwen2.5-VL-3B & $0.296$ & $0.422$ & $0.513$ & $-0.058^{*}$ & $-0.042^{*}$ & $-0.026^{*}$ \\
Qwen3-VL-8B & $0.496$ & $0.521$ & $0.563$ & $-0.011$ & $+0.007$ & $+0.024^{*}$ \\
Qwen3.5-9B & $0.463$ & $0.547$ & $0.570$ & $-0.003$ & $+0.014^{*}$ & $+0.031^{*}$ \\
\midrule
Mean, sig.\ $>$\,/\,$<$ & $0.425$ & $0.478$ & $0.553$ & $1\,/\,6$ & $3\,/\,5$ & $6\,/\,1$ \\
\bottomrule
\end{tabular}

\caption{\textbf{Per model, five main categories.} $\kappa_w$ under text parsing, at matched resolution ($P(\mathrm{yes})$ reduced to the parsed ratings' levels, \cref{sec:readout}) and under P(yes) scoring. $\Delta$: P(yes) scoring minus the three-expert reference, with the model re-mapped onto the same three experts' ratings (mean $0.543$, \cref{sec:anchor}), and minus EIF Small and Large, paired image bootstrap. $^{*}$ marks a $95\%$ CI excluding zero. Last row: mean, and models significantly above\,/\,below. All $40$ categories: \cref{tab:eif-full}.}
\label{tab:eif}
\end{table}

\textbf{Against expert--expert agreement.}
Under text parsing the eleven models average $0.425$, below expert--expert agreement. Two models, Qwen3-VL-8B and Gemma-4-12B-it, are significantly above it in a paired test, and six models are significantly below it (\cref{fig:forest}a). Under P(yes) scoring with the answer prefix, every model's interval lies above it, and the lowest lower bound is $0.496$. Each model's own gain is positive, so the mean gain is not carried by a few models. \emph{Anger} is the exception among the five: there the mean over models of $0.458$ falls below Anger's own expert--expert agreement, also $0.468$, with six models below (\cref{tab:percat}). A constant prediction scores $0.000$ and the benchmark's Random Baseline $0.008$ (\cref{tab:taxonomy,tab:published}).

\textbf{Against the benchmark's fine-tuned model.}
EIF is the benchmark's answer to the VLMs' failure: a model trained on EmoNet-Face for this task. Its two sizes trade places by category set: Small is ahead on the five main categories (\cref{tab:published}), Large on all $40$ ($0.267$ against $0.239$). We therefore compare against both with the same paired bootstrap. Without any training, three of the eleven VLMs are significantly above Small on the five categories and six are above Large, while five are below Small and one below Large. On all $40$, seven are significantly above Small and two above Large. GLM-4.6V-Flash and InternVL3\_5 are above both sizes on both sets.

\textbf{Against the three-expert reference.}
Under P(yes) scoring, GLM-4.6V-Flash is significantly above the stricter reference of \cref{sec:anchor}, InternVL3\_5 is above by point estimate, and six models are significantly below. Under text parsing all eleven are significantly below. Over all $40$ categories the three-expert reference is $0.263$, and under P(yes) scoring GLM-4.6V-Flash and InternVL3\_5 are significantly above it.

\textbf{Differences between models.}
\label{sec:spread}
A protocol that lifts every model could also erase the differences between them and so make the benchmark useless for ranking. It does not. The models differ less under P(yes) scoring (\cref{fig:forest}b), yet Holm correction over the $55$ model pairs still resolves $40$ of them, against $42$ under text parsing.

\section{What the Gain Depends On}
\label{sec:depends}

Our protocol makes choices the benchmark did not, and any of them could produce the gain instead of the readout. \Cref{tab:controls} changes one choice per row and re-scores the gain with the paired bootstrap. The gain stays positive under every choice.

\begin{table*}[t]
\centering\footnotesize\setlength{\tabcolsep}{4.5pt}
\begin{tabular}{lccccccc}
\toprule
Control & ref.\ & text & $P(\mathrm{yes})$ & $\Delta$ [95\% CI] & $P(\mathrm{yes})$ $>$ ref.\ & text $>$ ref.\ & $\Delta$, in-sample map \\
\midrule
As reported & $0.468$ & $0.425$ & $0.553$ & $+0.128$ $[+0.121, +0.133]$ & $11\,(11)/11$ & $3\,(2)/11$ & $+0.128$ \\
No answer prefix & $0.468$ & $0.415$ & $0.527$ & $+0.111$ $[+0.104, +0.118]$ & $7\,(7)/8$ & $2\,(2)/8$ & $+0.111$ \\
Coverage-matched images & $0.469$ & $0.425$ & $0.555$ & $+0.129$ $[+0.123, +0.135]$ & $11\,(11)/11$ & $3\,(2)/11$ & $+0.129$ \\
Benchmark's own prompt only & $0.468$ & $0.394$ & $0.553$ & $+0.159$ $[+0.152, +0.166]$ & $11\,(11)/11$ & $3\,(2)/11$ & $+0.160$ \\
Retry loop included & $0.468$ & $0.429$ & $0.553$ & $+0.124$ $[+0.118, +0.129]$ & $11\,(11)/11$ & $4\,(3)/11$ & $+0.124$ \\
In-sample mapping (upper bound) & $0.468$ & $0.426$ & $0.554$ & $+0.128$ $[+0.122, +0.134]$ & $11\,(11)/11$ & $3\,(2)/11$ & -- \\
Matched resolution & $0.469$ & $0.425$ & $0.478$ & $+0.052$ $[+0.046, +0.056]$ & $6\,(6)/11$ & -- & $+0.052$ \\
No mapping, raw scale & $0.468$ & $0.394$ & $0.415$ & $+0.021$ $[+0.014, +0.027]$ & $4\,(2)/11$ & $1\,(1)/11$ & -- \\
One expert through the mapping & $0.476$ & $0.425$ & $0.553$ & $+0.128$ $[+0.121, +0.133]$ & $11\,(11)/11$ & $3\,(2)/11$ & $+0.128$ \\
Three-expert panel, held-out expert & $0.563$ & $0.418$ & $0.543$ & $+0.125$ $[+0.119, +0.130]$ & $2\,(1)/11$ & $0\,(0)/11$ & $+0.125$ \\
Binary detection, no mapping & $0.388$ & $0.215$ & $0.380$ & $+0.164$ $[+0.158, +0.171]$ & $4\,(4)/11$ & $1\,(1)/11$ & -- \\
Binary detection, prevalence-matched & $0.388$ & $0.215$ & $0.458$ & $+0.243$ $[+0.233, +0.249]$ & $11\,(11)/11$ & $1\,(1)/11$ & -- \\
Mean of 16 sampled answers & $0.468$ & $0.424$ & $0.553$ & $+0.129$ $[+0.122, +0.137]$ & $11\,(11)/11$ & $3\,(3)/11$ & $+0.129$ \\
Graded digit, argmax (500) & $0.454$ & $0.472$ & $0.527$ & $+0.055$ $[+0.044, +0.064]$ & $11\,(11)/11$ & $7\,(5)/11$ & $+0.054$ \\
Graded digit, expected (500) & $0.454$ & $0.521$ & $0.527$ & $+0.006$ $[-0.002, +0.013]$ & $11\,(11)/11$ & $11\,(8)/11$ & $+0.006$ \\
\bottomrule
\end{tabular}

\caption{Controls, five main categories. text: text parsing. $\Delta$: P(yes) scoring minus text parsing, mean over the models and images each row admits (graded digit: $500$-image subset), paired image bootstrap. $P(\mathrm{yes}) >$ ref., text $>$ ref.: models above the row's reference by point estimate, significantly in parentheses. Last column: $\Delta$ under the in-sample mapping. In the graded-digit rows, \emph{text} holds the graded-digit score.}
\label{tab:controls}
\end{table*}

\textbf{Choices that could favor P(yes) scoring.}
\emph{No answer prefix}: only eight models place their first-token mass on \texttt{yes} or \texttt{no} without the prefix, and seven of them still clear expert--expert agreement (\cref{sec:prefill}). \emph{Coverage-matched images}: text parsing loses the images whose reply it cannot parse, so we restrict both conditions to each model's parsed images; the gain is unchanged (\cref{sec:app-coverage}). \emph{In-sample mapping (upper bound)}: letting the mapping see the ratings it is scored against moves each condition by at most $0.001$. \emph{One expert through the mapping}: to check that the mapping does not favor models over people, we pass one expert through it like a model; that expert agrees with the other three at $0.476$, and all eleven models stay significantly above. \emph{No mapping, raw scale}: with parsed ratings as emitted and P(yes) as $\mathrm{round}(7\,P(\mathrm{yes}))$, both conditions fall, and the gain shrinks to $+0.021$ but stays positive. \emph{Mean of 16 sampled answers}: averaging several sampled replies gives text parsing a graded score of its own, so we average sixteen replies per image on all eleven models and $2{,}500$ images under model-card decoding. The average scores $0.086$ above one sampled reply, matches the reported, mostly greedy, text score ($-0.002$, n.s.) and stays $0.129$ below P(yes) scoring.\footnote{MiniCPM-V-4.6 ran this control at $150{,}000$ pixels, because it fails on multi-tile images in this setup.}

\textbf{Choices that favor text parsing.}
\emph{Benchmark's own prompt only}: without our second prompt, text parsing falls to $0.394$ (\cref{sec:settings}). \emph{Retry loop included}: Gemma-3-12b reaches $0.491$, one more text-parsing model is significantly above the reference, and the gain barely moves (\cref{sec:retry}).

\textbf{Other references and decompositions.}
\emph{Three-expert panel, held-out expert}: the gain is similar ($+0.125$), with both scores lower (\cref{sec:results}). \emph{Matched resolution} and the two \emph{binary detection} rows split the gain into its sources (\cref{sec:readout}). Other tie rules and other target rating distributions for the mapping move the gain between $+0.107$ and $+0.149$ (\cref{tab:tierule}). Training does not substitute for the readout: LoRA fine-tuning on $19{,}451$ yes/no queries from EmoNet-Face-Binary leaves P(yes) scoring statistically unchanged on six of eight models (\cref{sec:app-training}).

\subsection{Ordering and resolution}
\label{sec:readout}
We split the gain two ways: by what the score carries (ordering and resolution, here) and by what in the protocol produces it (question format and readout, below). A parsed rating and $P(\mathrm{yes})$ differ in which images a model ranks higher (ordering) and in how many distinct values it emits (resolution). To separate the two, we requantize each model's $P(\mathrm{yes})$ to that model's own parsed-rating histogram, per category, on its parsed images (reference recomputed on those images, $0.469$). This \emph{matched resolution} keeps the parsed ratings' levels, counts and ties, and differs only in which image receives which value. Better ordering is worth $+0.052$ $[+0.046, +0.056]$, about $40\%$ $[37, 42]$ of the gain. It is significantly positive for seven of eleven models and significantly negative for none. mAP moves the same way ($0.561$, $0.590$ and $0.773$ for text parsing, matched resolution and P(yes) scoring). The remaining $60\%$ is resolution.

\textbf{Detection carries much of the gain.}
Cohen's $\kappa$ on present versus absent isolates detection. Without any mapping, reading text as present above $0$ and $P(\mathrm{yes})$ as present above $0.5$, P(yes) scoring detects at $0.380$ against $0.215$, close to the $0.388$ at which two experts agree. Thresholded at the experts' rate of presence, which only a continuous score can be cut at, it reaches $0.458$, an upper bound.

\textbf{Question format and readout.}
\label{sec:format}
Text parsing has no clean continuous counterpart to $P(\mathrm{yes})$, because the probability of a category name is split across the many ways of writing it~\cite{holtzman2021surface}. The two graded-digit rows of \cref{tab:controls} ask the same $40$ one-category queries as P(yes) scoring, but for a rating from $0$ to $7$ in one digit, on a $500$-image subset (reference $0.454$) (\cref{sec:app-rate}). \emph{Graded digit, argmax}: the digit a greedy decode prints scores $0.472$, above expert--expert agreement by point estimate and $+0.059$ $[+0.046, +0.074]$ above parsed text on these images. \emph{Graded digit, expected}: reading the probability-weighted mean of the eight digits adds $+0.048$ $[+0.042, +0.056]$, significant for all eleven models, and P(yes) scoring adds only $+0.006$ $[-0.002, +0.013]$ beyond that. The gap between parsed text and P(yes) scoring thus splits into two parts of similar size, asking one category per query and reading the probability. Asking yes/no instead of for a digit ($0$--$7$) therefore adds nothing. Read as a hard yes/no instead of a probability, by thresholding $P(\mathrm{yes})$ at $0.5$, the answers form two tied blocks under the mapping and lose to parsed text on all images ($0.362$ against $0.425$), because two levels carry no intensity (\cref{sec:app-readout}).

\subsection{Answer prefix}
\label{sec:prefill}
The answer prefix is the one asymmetry between the conditions that is not intrinsic to the scoring rule. Clearing expert--expert agreement depends on it for four of eleven models. For GLM, MiMo and Qwen3.5 it is a precondition: without it they place almost no first-token mass on \texttt{yes} or \texttt{no}, so there is no answer to score. The fourth, Gemma-3-4b-it, is the case the answer-mass check misses. Its answer mass is $0.965$ without the prefix, yet the prefix is worth $0.139$ $\kappa_w$ to it, where none of the other seven models that answer without it gains more than $0.009$. Without the prefix it falls below expert--expert agreement (\cref{tab:prefill}).

\subsection{Phrasing and category set}
\label{sec:robustness}
\label{sec:taxonomy}
The query wording and the main categories are our choices, so we vary both. Three phrasings of the query (\cref{sec:app-phrasings}) across eleven models give $33$ model--phrasing pairs. Every pair is above expert--expert agreement by point estimate (minimum $0.481$), and $31$ of $33$ significantly. With the answer prefix, every model clears expert--expert agreement on the five main categories, on the other $35$ and on all $40$ (\cref{tab:taxonomy}). On all $40$, text parsing averages $0.135$ and is significantly below the expert--expert $0.202$ for ten of eleven models, while P(yes) scoring is significantly above it for all eleven. Expert agreement decays smoothly across the taxonomy, with no gap at the main categories' threshold (\cref{fig:alpha}): $11$ of $40$ categories fall below $\alpha = 0.10$, and three below zero.

\section{Real Photographs}
\label{sec:replication}

All results so far are on synthetic portraits. We therefore test whether the gain holds on real photographs, with FACES~\cite{ebner2010faces}: $2{,}052$ photographs of $171$ people posing six expressions (anger, disgust, fear, happiness, neutral and sadness). One expression is correct per photograph, so FACES is a six-way forced choice.

FACES distributes aggregated validation ratings, not the per-rater judgments an expert--expert reference needs. We therefore score against the posed expression and use neither that reference nor the quantile mapping. Text parsing asks for exactly one of the six expressions and parses the word it returns. P(yes) scoring asks the yes/no question for each expression and predicts the most probable. The metric is accuracy (chance $1/6$). Intervals resample people, not images, since each person contributes twelve photographs.

\textbf{On photographs the gain is small.}
Over all eleven models the mean accuracy gain is $+0.030$ $[+0.023, +0.036]$. Six models gain significantly, three are inconclusive, and two reverse, Gemma-3-12b-it ($-0.063$) and MiMo-VL-7B-RL ($-0.020$) (\cref{tab:faces}). MiMo is the one text-parsing run of eleven that fails the truncation check (\cref{sec:gate}). It is marked, not excluded, and leaving it out would raise the mean to $+0.035$ $[+0.028, +0.042]$. EmoNet-Face has no neutral category, so we also rerun both conditions with neutral removed from the prompt and the photographs. Then the ten models that pass the truncation check give two significant gains, five inconclusive results and three significant reversals, mean $-0.018$ (\cref{sec:app-fiveclass}). Both FACES variants differ from EmoNet-Face-HQ: one expression is correct, and the prediction is the most probable class. The graded resolution that carries much of the gain there is not scored here. FACES therefore shows that P(yes) scoring runs unchanged on real photographs and a forced-choice task, not how the graded gain transfers to them. Macro-F1, per-model detail and controls are in \cref{sec:app-faces}.

\section{Limitations}
\label{sec:limitations}

\textbf{Synthetic faces, one benchmark family.}
The graded result rests on generated, cropped single-face portraits. In-the-wild affect, which often needs scene context and several labels per person~\cite{kosti2017emotic}, remains untested. FACES is single-label, so it does not test the graded part of the gain on real photographs, and its five-class rerun reverses the sign (\cref{sec:replication}).

\textbf{Mapping and category cut.}
Every mapped score uses a mapping fitted to the experts' rating distribution. It is fitted on held-out images, so no score uses the ratings it is compared with, but applying the protocol to a new dataset requires expert-rated images per category. Without the mapping, P(yes) scoring is significantly above expert--expert agreement for two models (\cref{tab:controls}). The threshold for the five main categories is close (\cref{sec:data}), which is why every result is also given on all $40$.

\textbf{Answer prefix.}
Four models depend on the answer prefix (\cref{sec:prefill}). Three give no answer without it, and Gemma-3-4b-it then falls below expert--expert agreement.

\section{Conclusion}
\label{sec:conclusion}

EmoNet-Face-HQ makes expert agreement on fine-grained facial emotion measurable, and its protocol understates the models it scores. That agreement is itself modest: over all $40$ categories two experts reach $\kappa_w = 0.202$, and only five categories reach $\alpha \ge 0.30$, because many categories are close in meaning or hard to read from a still face. Any score on the benchmark should therefore be read against a human reference scored the same way, per category. Scored by $P(\mathrm{yes})$ with an answer prefix and a held-out quantile mapping, all eleven open-weight VLMs agree with an individual expert at least as well as a second expert does on the five most reliable categories. Only one is significantly above the stricter three-expert reference. The gain is part ordering, part resolution, and lies in detecting whether an emotion is present. It is small on real posed photographs, and it concerns observer ratings of displayed expression, not felt emotion.

\textbf{Implications.} A benchmark measures a model through a procedure, and when the procedure contributes more to the score than the model does, the number describes the procedure. For facial affect analysis, VLM results scored from parsed text, here and in comparable evaluations~\cite{lian2024gpt4v}, likely understate what the models perceive. A pre-trained VLM read by probability is the baseline a fine-tuned model should beat before it is called necessary. Other graded face labels, such as action-unit intensity or valence and arousal, admit the same readout and should be tested with it. Beyond faces, any evaluation in which a model rates on a graded scale or serves as an annotator measures its protocol along with the model: here the prompt alone moves one model from chance to near expert agreement.

For benchmarks that score VLMs on graded affect labels we recommend three practices:
\begin{enumerate}
\item Report the prompt and the scoring rule as part of the protocol, and do not fix how many labels a model must return.
\item Score expert--expert agreement and trivial predictors through the same pipeline as the models, per category, beside each category's agreement and prevalence.
\item Ask one category at a time and, where answer probabilities are available, report the probability-based score beside the parsed-text score, mapped onto the label scale on held-out images.
\end{enumerate}

\section*{Acknowledgment}
Part of the experiments ran on the LiCCA HPC cluster of the University of Augsburg, co-funded by the Deutsche For\-schungs\-ge\-mein\-schaft (DFG, German Research Foundation)~-- Project-ID~499211671.

{
    \small
    \bibliographystyle{ieee}
    \bibliography{main}
}

\end{document}